\documentclass[preprint,review, 12pt]{elsarticle}

\usepackage{amssymb}
\usepackage{amsmath}

\usepackage{graphicx}
\usepackage{amsmath}
\usepackage{amssymb}
\usepackage{lipsum}
\usepackage{multirow}
\usepackage{multicol}
\usepackage{makecell}
\usepackage{subfigure}
\usepackage[table]{xcolor}
\usepackage{float}
\usepackage{algorithmic}
\usepackage{algorithm}
\restylefloat{table}

\usepackage{booktabs}       
\usepackage{multirow}       
\usepackage{graphicx}       
\usepackage{caption}        
\usepackage{tabularx}
\usepackage[colorlinks,linkcolor=blue,anchorcolor=black,citecolor=blue]{hyperref}
\usepackage{tcolorbox}
\usepackage{colortbl}

\usepackage[capitalize]{cleveref}

\renewcommand{\arraystretch}{1.5}

\definecolor{Blue}{gray}{0.85}
\newcolumntype{b}{>{\columncolor{Blue}}c}

\journal{Reliability Engineering \& System Safety}

\makeatletter
\def\ps@pprintTitle{%
  \let\@oddhead\@empty
  \let\@evenhead\@empty
  \def\@oddfoot{\reset@font\hfil\thepage\hfil}
  \let\@evenfoot\@oddfoot
}
\makeatother

\begin{document}

\begin{frontmatter}



\title{Multimodal Learning for Arcing Detection in Pantograph-Catenary Systems}



\affiliation[inst1]{organization={ELLIS Institute Finland and Tampere University}, city={Espoo}, country={Finland}}
\affiliation[inst2]{organization={Chair of Structural Mechanics \& Monitoring, ETH Z\"urich}, city={Z\"urich}, country={Switzerland}}
\affiliation[inst3]{organization={Intelligent Maintenance and Operations Systems, EPFL}, city={Lausanne}, country={Switzerland}}
\author[inst1,inst2]{Hao Dong}
\author[inst2]{Eleni Chatzi}
\author[inst3]{Olga Fink}

\begin{abstract}
The pantograph-catenary interface is essential for ensuring uninterrupted and reliable power delivery in electrified rail systems. However, electrical arcing at this interface poses serious risks, including accelerated wear of contact components, degraded system performance, and potential service disruptions. Detecting arcing events at the pantograph-catenary interface is challenging due to their transient nature, noisy operating environment, data scarcity, and the difficulty of distinguishing arcs from other similar transient phenomena. To address these challenges, we propose a novel multimodal framework that combines high-resolution image data with force measurements to more accurately and robustly detect arcing events. 
First, we construct two arcing detection datasets comprising synchronized visual and force measurements. One dataset is built from data provided by the Swiss Federal Railways (SBB), and the other is derived from publicly available videos of arcing events in different railway systems and synthetic force data that mimic the characteristics observed in the real dataset.
Leveraging these datasets, we propose MultiDeepSAD, an extension of the DeepSAD algorithm for multiple modalities with a new loss formulation. Additionally, we introduce tailored pseudo-anomaly generation techniques specific to each data type, such as synthetic arc-like artifacts in images and simulated force irregularities, to augment training data and improve the discriminative ability of the model. Through extensive experiments and ablation studies, we demonstrate that our framework significantly outperforms baseline approaches, exhibiting enhanced sensitivity to real arcing events even under domain shifts and limited availability of real arcing observations. To the best of our knowledge, this is the first method and publicly available dataset that integrates image and force data for pantograph-catenary arcing event detection. The proposed framework offers a practical solution for real-time monitoring of arcing events in pantograph-catenary systems, ultimately contributing to safer and more reliable railway operations. Our source code and dataset are at \href{https://github.com/EPFL-IMOS/Multimodal-Arcing}{https://github.com/EPFL-IMOS/Multimodal-Arcing}.
\end{abstract}







\end{frontmatter}

\section{Introduction}
\label{introduction}

The pantograph-catenary system serves as the critical interface for power transmission to electric trains, ensuring continuous and reliable operation across electrified rail networks~\cite{diyang2024impactability,diyang2024reasoning,mo2025knowledge,wang2023data}. Reliable electrical contact at this interface is essential for the safe and efficient functioning of railway systems, supplying uninterrupted power for both traction and onboard services~\cite{bruni2018pantograph,xinyuan2025causal,fink2026physics,wang2024uncertainty,fink2025physics}. Any disruption at this interface can lead to operational delays, diminished performance, and potential damage to the complex electrical equipment involved. The growing global demand for efficient, reliable and modern rail transport, spanning both passenger and freight services, highlights the need for dependable power delivery, thereby increasing the importance of detecting and mitigating anomalies such as electrical arcing in the pantograph-catenary system~\cite{aydin2015anomaly,chen2022high}. As operational demands on rail systems intensify, whether due to increased speed, higher frequency of service, or heavier loads, the dynamic interaction between the pantograph and the catenary becomes more complex, increasing the risk of contact loss and subsequent arcing. Therefore, effective arcing detection and assessment of their contribution to degradation is increasingly important for maintaining reliable operation across different service conditions~\cite{gao2018pantograph,liu2023novel}. 

Electrical arcing in pantograph-catenary systems is a sporadic phenomenon that stems from insufficient contact between the pantograph contact strip and the overhead catenary wire~\cite{karaduman2017deep}. Poor contact conditions can be attributed to a number of factors, including inherent vibrations of the pantograph, irregularities in the geometry of the contact line, the presence of hard spots on the contact surfaces, and the passage of the train over electrical neutral zones~\cite{gao2018pantograph}. Furthermore, environmental effects, such as the formation of ice on contact surfaces during colder periods, can also contribute to the initiation of arcing. The generation of an electric arc is accompanied by the production of intense heat, which, over time, leads to the degradation and erosion of both the sliding plate of the pantograph and the catenary wire itself~\cite{wang2013review}. Without continuous monitoring and timely mitigation, repeated arcing events can accumulate damage to both the pantograph and the catenary infrastructure, requiring costly maintenance interventions and potentially resulting in significant disruptions to railway services~\cite{karaduman2017deep}. Beyond direct physical damage, arcing events also generate electromagnetic disturbances that can pose a threat to the electromagnetic compatibility and safety of the trains and may interfere with vital communication systems utilized for railway signaling and control. The high temperatures generated by arcing accelerate contact-surface wear, progressively degrading contact quality and increasing the likelihood of further arcing without sustained monitoring and maintenance. Arcing is, thus, not merely a sign of poor contact, but a destructive process that actively accelerates deterioration, highlighting the importance of continuous monitoring and assessment of cumulative arcing activity to minimize its long-term impact on railway infrastructure and operational reliability.

\begin{figure}[t!]
 \centering \includegraphics[width=0.65\linewidth]{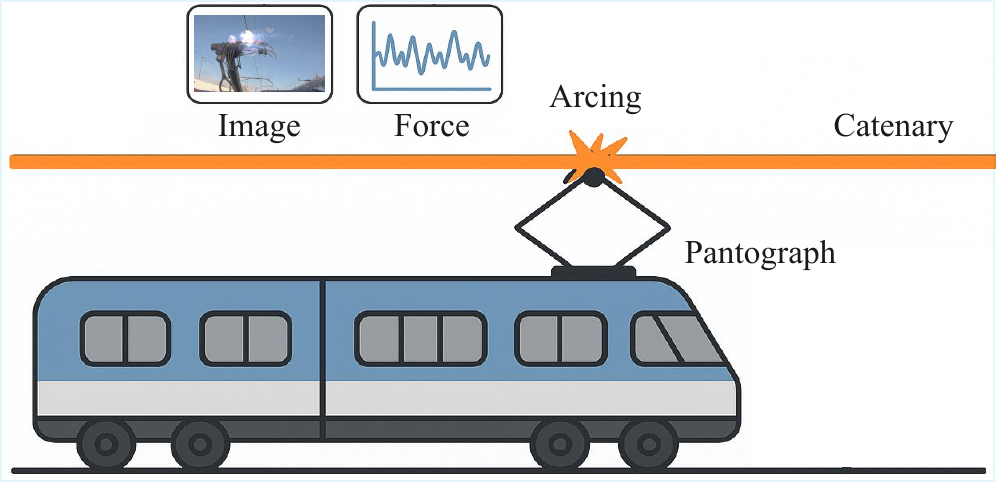}
\vspace{-0.2cm}
\caption{Proposed solution for arcing detection in pantograph-catenary systems with image and force measurements. }
\label{fig:intro}
\end{figure}

Traditional methods for detecting arcing in pantograph-catenary systems have primarily relied on analyzing electrical signals, such as voltage and current, within the power supply system~\cite{barmada2014arc,huang2018cluster,yang2025arc}. However, arcing-induced signatures can often be masked by noise and other electrical phenomena in these signals, making detection a challenging task that requires advanced feature extraction and algorithm design~\cite{li2025theoretical}. 
In response to the limitations of traditional electrical signal-based methods, growing interest has turned to leveraging image processing and computer vision techniques for arcing detection~\cite{yan2025novel}. These methods provide a non-contact means of monitoring the pantograph-catenary interface, enabling the identification of the distinctive visual signatures associated with electrical arcing. Complementary to visual methods, force and accelerometer data offer valuable insights into the mechanical interaction between the pantograph and the catenary~\cite{tan2024defect}. Such sensors can detect events such as contact loss or excessive vibration, which often precede or accompany the occurrence of electrical arcs. 

In this research, we propose a unified, multimodal framework that integrates image and contact force measurements to achieve  more robust and accurate arcing detection (\cref{fig:intro}). By leveraging the complementary information of both visual and mechanical sensing, our approach offers a reliable solution for monitoring pantograph-catenary system health, particularly under complex operational conditions. 
Due to the absence of public datasets for railway arcing detection, we first create two arcing detection datasets comprising synchronized visual and force measurements. One dataset is derived from data provided by Swiss Federal Railways (SBB), while the other is collected from publicly available internet videos and synthetic force data. The second dataset will be publicly released to facilitate future research in this domain.
We then propose MultiDeepSAD, an extension of DeepSAD~\cite{ruff2019deep} adapted to multimodal settings with \textit{a new loss formulation}, and develop modality-specific \textit{pseudo-anomaly generation strategies} for both image and force inputs. These pseudo-anomalies are then used to train MultiDeepSAD, enabling improved arcing detection performance. Extensive experiments and ablation studies demonstrate the robustness of our framework across varying numbers of real anomaly samples for pseudo-anomaly generation and domain shifts. Our contributions can be summarized as follows:
\begin{itemize}
\setlength\itemsep{0em}
\item We propose MultiDeepSAD, a universal framework for multimodal arcing detection in pantograph-catenary systems, incorporating both a new training objective and modality-specific pseudo-anomaly generation strategies.
\item We release a new multimodal dataset that includes synchronized visual and force measurements, which can serve as a challenging benchmark for future research on arcing detection in pantograph-catenary systems.
\item Extensive experiments and ablation studies across both datasets consistently demonstrate the effectiveness and robustness of the proposed framework.
\end{itemize}

\section{Related Work}

\noindent\textbf{Current/Voltage-based Arcing Detection.} Early arcing detection methods primarily relied on handcrafted features~\cite{mariscotti2023electrical}, including basic time-domain statistical analysis and frequency-domain methods like the Fast Fourier Transform~\cite{en14020288}. To better analyze the non-stationary nature of arc transients, more advanced time-frequency techniques such as the Wavelet Transform and the Hilbert-Huang Transform~\cite{barmada2011use} were introduced, enabling simultaneous localization of arc events in both time and frequency. The field has increasingly transitioned towards machine learning-based methods, first employing models like Support Vector Machines~\cite{barmada2014arc} that rely on handcrafted features extracted from these signals. Besides, other methods that combine the Discrete Wavelet Transform with deep neural networks have become prominent~\cite{yu2019identification}. This approach leverages the Wavelet transform to decompose the signal, allowing the neural network to automatically learn complex fault characteristics from the resulting time-frequency data, achieving high recognition accuracy.

\noindent\textbf{Image-based Arcing Detection.}
Instead of relying on current or voltage measurements—which are often unavailable, noisy, and highly dependent on sensor placement and operating conditions—recent research has increasingly focused on visual inputs, as arcing typically manifests as distinctive light emission and sparks that can be captured by onboard cameras and processed robustly with modern vision models~\cite{huang2018cluster,liu2024pantograph,gao2020automatic,li2024virtual}. High-speed cameras were used to capture the transient nature of arcing, with subsequent image processing techniques developed to isolate and characterize arc events~\cite{aydin2015anomaly}. More recent advances have leveraged machine learning algorithms to automate the detection and classification of arcing events, thereby improving both accuracy and robustness~\cite{huang2019arc,quan2023arcmask,liu2023novel}. For example, Quan et al~\cite{quan2024arcse} introduce a dual-branch semantic segmentation model designed for robust pantograph-catenary arcing detection. The architecture features a Semantic Feature Branch that captures multi-scale global context and a parallel Detail Feature Branch that preserves high-resolution information to accurately segment small arcing events. A novel Feature Enhancement Mechanism subsequently fuses the outputs and employs a learnable visual center module to increase the feature discriminability between arcing and complex backgrounds, thereby improving model robustness. Building on the idea of multiscale representation, Liu et al.~\cite{liu2023novel} utilize a Swin Transformer encoder to capture global context and a custom down-top multi-scale CNN decoder that fuses features from different levels to handle various arcing sizes. A key contribution is the Arc Feature Augmentation module, which leverages the inherent high luminosity of arcing by combining global max and threshold features to selectively amplify the arcing pixel representation, significantly improving detection in complex scenes. In a complementary direction, Yan et al.~\cite{yan2025novel} address the challenge of detecting arcs with varying sizes and shapes by integrating a guided anchor mechanism that dynamically adjusts anchor generation based on arc characteristics, improving both efficiency and localization accuracy.

\noindent\textbf{Force/Accelerometer-based Arcing Detection.}
Complementary to image-based methods, sensor data from force sensors and accelerometers have been used to capture the mechanical vibrations and dynamic loads associated with arcing~\cite{wu2019diagnosis,elia2006condition,gregori2023assessment}. Such approaches take advantage of the fact that electrical arcing can produce detectable force disturbances and vibrational signatures in the pantograph-catenary system~\cite{hong2024non}. Researchers have developed signal processing pipelines to extract characteristic patterns from these sensors, which can be correlated with arc events~\cite{pappalardo2016contact,gao2017detection}.
For example, in the preprocessing pipeline of Gregori et al.~\cite{gregori2023assessment}, raw acceleration signals are resampled to a common spatial domain to achieve speed-invariance, segmented into samples, and then standardized before being used as input for the networks.

\noindent\textbf{Multimodal Arcing Detection.}
The integration of image data with other modalities is an emerging trend that addresses the limitations of single-modality approaches. By integrating information from visual and complementary sensors such as infrared and audio, it has become possible to mitigate false positives caused by environmental noise and sensor-specific artifacts. For example, Huang et al.~\cite{huang2020arc} fuse RGB and infrared images using a convolutional neural network (CNN) for arc detection in visible-light images, a threshold method for infrared images, and a CNN-based environment perception model to dynamically adjust weights for each modality before fusing the results using evidence theory. Yan et al~\cite{yan2025research} propose a multimodal arc detection network that pretrains a Denoising Diffusion Probabilistic Model~\cite{ho2020denoising} on unlabeled infrared and visible light images for feature extraction and then fine-tunes a decoder that uses audio signals as semantic prompts to improve visual arc detection with limited data. Despite advancements in the field, the fusion of image and force data has been largely overlooked for arcing detection. 

\noindent\textbf{Pseudo-anomaly Generation.} Since pantograph arcing events are rare, safety-critical, and often weakly labeled in real operational data, several recent works motivate framing arc detection as an anomaly detection problem, where pseudo-anomaly generation can provide additional supervision. Prior work on pseudo-anomaly generation has demonstrated that synthesizing realistic yet controllably abnormal samples can substantially enhance the training of anomaly detectors by providing a richer supervisory signal in the absence of labeled anomalies~\cite{dong2024multiood,liu2025fm,nejjar2024recall}. Early approaches leveraged simple perturbations, such as geometric transformations~\cite{golan2018deep} or patch‐level modifications~\cite{li2021cutpaste}, to create anomaly examples for one‐class classifiers, while more recent methods employ deep generative models to produce high-fidelity anomalies that better mimic the complexity of real deviations~\cite{sun2025unseen,schlegl2019f}. Furthermore, recent studies have explored semi-supervised paradigms, in which a small set of labeled anomalies guides the generation process~\cite{dong2023nngmix}, as well as adaptive techniques that dynamically adjust the difficulty of generated anomalies throughout training~\cite{lappas2024dynamic}. However, the intersection of multimodal learning and arcing event detection remains largely unexplored in current literature.

\section{The Creation of Multimodal Arcing Detection Dataset}

Due to the limited availability of public datasets for railway arcing detection, we construct two arcing detection datasets to support our experiments.

\begin{figure}[t!]
 \centering \includegraphics[width=\linewidth]{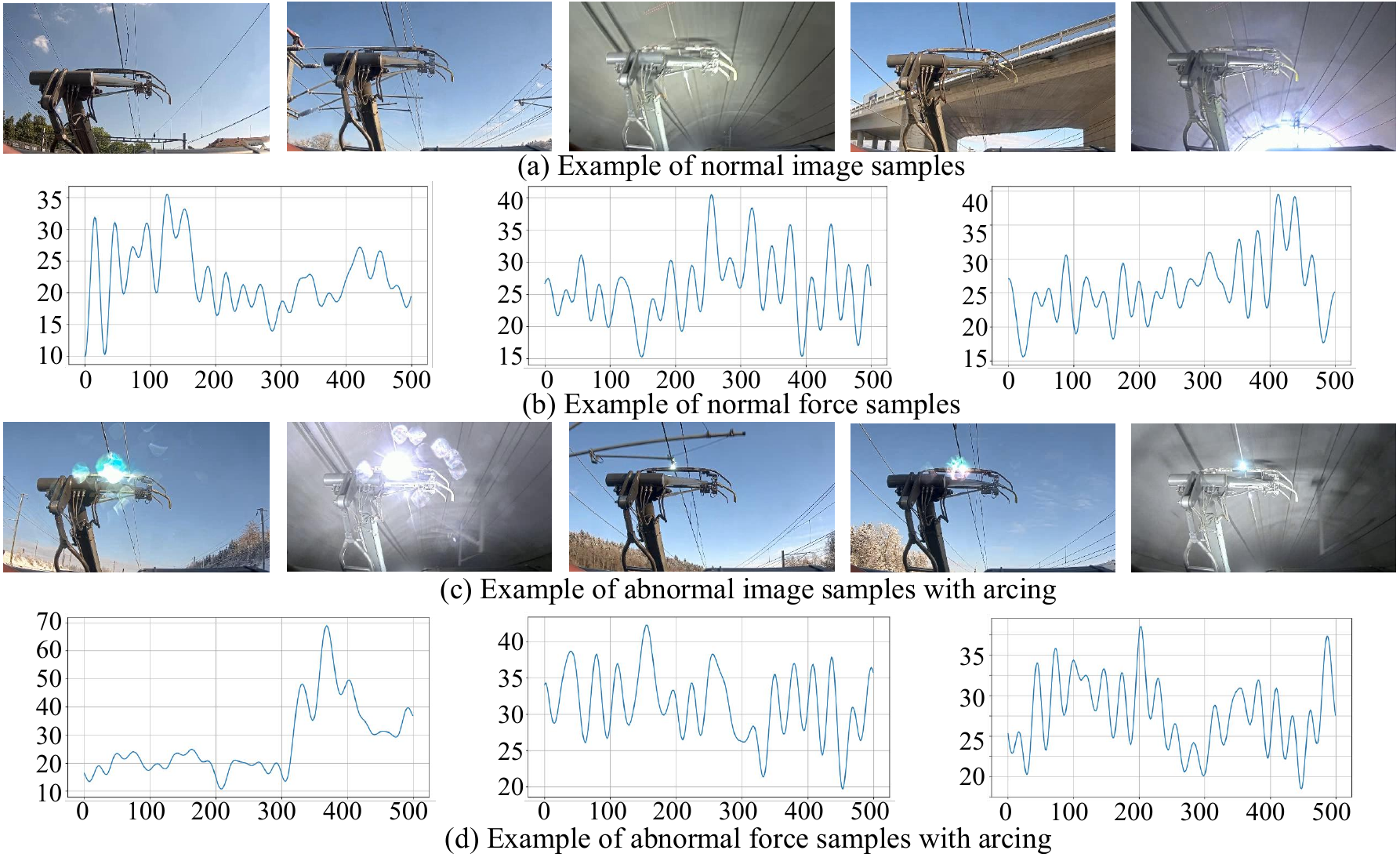}
\vspace{-1.0cm}
\caption{Representative samples from the SBB-AD dataset, consisting of time-synchronized pantograph camera images and corresponding contact-force measurements provided by SBB.}
\label{fig:datasets}
\end{figure}

\subsection{Multimodal Arcing Detection Dataset from SBB Data}
The first dataset, termed the SBB Arcing Detection Dataset (SBB-AD), is created using the data provided by our industrial partner SBB. This dataset includes synchronized image and force sensor data (\cref{fig:datasets}), captured from an SBB diagnostic train operating in Switzerland. The data acquisition setup involved a camera focused on the pantograph-catenary system and a co-located force sensor measuring their contact force. The image data is recorded at 60 Hz, while the force data is sampled at 500 Hz. Collection campaigns were conducted in summer (August 8, 2022) and winter (January 19, 2023) to capture diverse acquisition conditions and background settings, including instances of tunnel passage, covering over 200 km in total. The dataset consists of 3,107 normal samples for training and a balanced test set of 328 samples, comprising 164 normal, 164 arcing anomaly instances. {To generate these samples, recorded videos were segmented into one-second intervals, with all arcing events carefully identified and labeled. For each segment containing an arcing event, a representative frame capturing the arc and its corresponding one-second force data were selected. 
For segments without arcing, one frame was randomly sampled along with its corresponding force data to represent normal operation. }
Due to copyright restrictions imposed by SBB, this dataset is private and used solely for experimental purposes. To facilitate future research in this domain and benefit the community, we additionally construct a second dataset that is fully open-sourced.

\begin{figure}[t!]
 \centering \includegraphics[width=\linewidth]{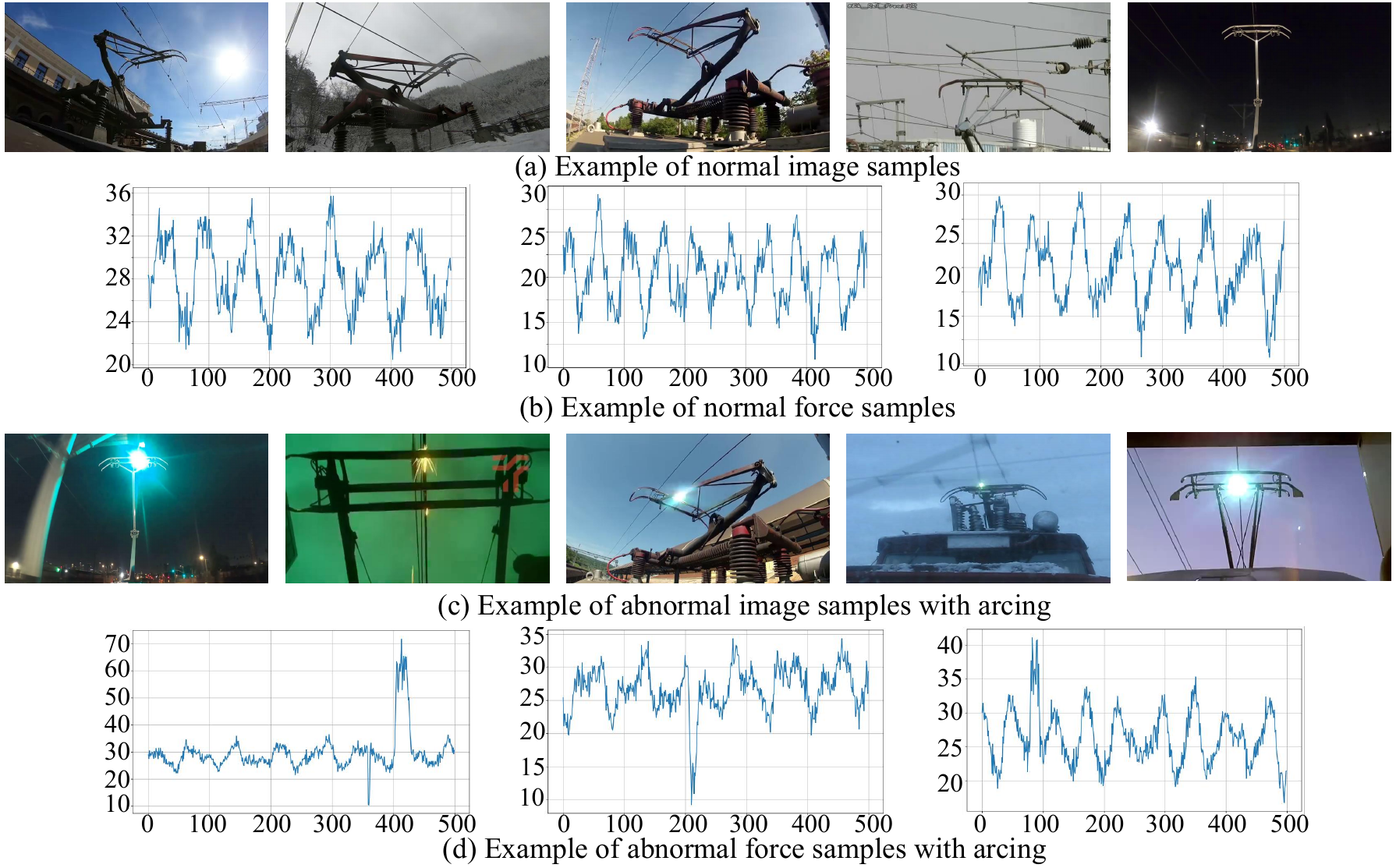}
\vspace{-1.0cm}
\caption{Examples from the Open-AD dataset, constructed from internet and simulation data. }
\label{fig:datasets1}
\end{figure}

\subsection{Multimodal Arcing Detection Dataset from Internet and Simulation Data}
The second dataset, termed the Open Arcing Detection Dataset (Open-AD), is constructed by combining internet-sourced video data with synthetic force data (\cref{fig:datasets1}). For the visual modality, we collect YouTube videos of pantograph-catenary systems that cover both normal operation and arcing events. For the force modality, we generate synthetic force signals under normal and abnormal conditions using hand-crafted signal properties designed to approximate real-world operating characteristics. We then extract image frames from the videos and associate each frame with a simulated force signal: normal force signals are paired with normal images, whereas abnormal force signals are paired with images that contain arcing events.

\subsubsection{Internet-sourced Data for Image Measurements}
For the image measurements, we collect videos of pantograph-catenary systems from YouTube, covering both normal operation and arcing events. In total, we curate $23$ videos spanning diverse weather and lighting conditions (e.g., sunny and snowy; daytime and nighttime), multiple pantograph-catenary configurations, and recordings from different countries, as shown in \cref{fig:datasets1} (a) and (c). We extract all frames from the videos and manually annotate those containing arcing events. Overall, the resulting dataset contains $3,800$ normal samples for training and a balanced test set of $484$ samples, comprising $238$ normal samples and $246$ arcing anomaly instances.

\subsubsection{Synthetic Data Generation for Force Measurements}

Because the internet videos do not provide corresponding force measurements, we generate synthetic force signals that mimic pantograph-catenary interactions under both normal and abnormal conditions. Each measurement consists of $500$ samples over a $1$-second interval, corresponding to a sampling rate of $500$ Hz, as shown in \cref{fig:datasets1} (b) and (d). 

\textbf{Normal condition samples} incorporate multiple sources of variability following the mathematical formulation:
\begin{equation}
F_{\text{normal}}(t) = F_{\text{mean}} + \sum_{i=1}^{3} A_i \cdot \sin(2\pi f_i t + \phi_i) + \eta_{\text{meas}}(t) + \eta_{\text{irreg}}(t) + D(t) + T(t) + V(t),
\label{eq:1}
\end{equation}
where $F_{\text{mean}} \sim \mathcal{U}(20, 30)$ N is the randomized mean contact force; the oscillatory components have frequencies $f_1 \sim \mathcal{U}(6, 10)$ Hz, $f_2 \sim \mathcal{U}(10, 14)$ Hz, $f_3 \sim \mathcal{U}(18, 25)$ Hz with corresponding amplitudes $A_1 \sim \mathcal{U}(2.5, 4.5)$ N, $A_2 \sim \mathcal{U}(1.2, 2.5)$ N, $A_3 \sim \mathcal{U}(0.6, 1.6)$ N, and random phase offsets $\phi_i \sim \mathcal{U}(0, 2\pi)$; $\eta_{\text{meas}} \sim \mathcal{N}(0, 1.0)$ N and $\eta_{\text{irreg}} \sim \mathcal{N}(0, 0.6)$ N represent Gaussian measurement noise and wire irregularities; $D(t) = d \cdot t$ with $d \sim \mathcal{U}(-1.5, 1.5)$ N represents linear temporal drift; $T(t)$ represents transient disturbances (30\% probability, Gaussian-shaped, $\pm 3$ N amplitude, $10-40$ ms duration); and $V(t)$ represents brief sinusoidal force variations ($50$\% probability, $\pm 2$ N amplitude, $16-50$ ms duration) that mimic subtle arcing signatures. 
{$T(t)$ models high-amplitude, short-duration transients caused by routine operational events, such as the pantograph passing over catenary hangers, track switches, or minor aerodynamic gusts. $V(t)$ represents low-amplitude, longer-duration mechanical variations, simulating friction changes or subtle structural resonances that do not result in contact loss or arcing.}
The mathematical formulation models the physical reality of pantograph-catenary contact force as a superposition of \textit{deterministic mechanical oscillations and stochastic disturbances}. These normal variations are constructed to create ambiguity with abnormal events, ensuring classification requires detailed signal analysis rather than simple thresholds.

\textbf{Abnormal condition samples} contain three types of arcing events with varying severity: (1) \textit{loss of contact arcing} ($35$\% probability) causing $30-90$\% force drops with electrical noise and contact impulses; (2) \textit{excessive contact force arcing} ($30$\% probability) producing $40-150$\% force increases with mechanical vibrations and oscillatory decay; and (3) \textit{subtle micro-arcing} (35\% probability) with minimal force perturbation ($\pm 5-20$\%) but elevated high-frequency noise ($\sigma = 0.8$ N). Each abnormal sample contains $1-3$ randomly positioned arcing events ($5-30$ samples duration), except $20$\% of abnormal samples that contain no discrete arcing events but only altered burst noise patterns ($\sigma = 1.2$ N over $20-60$ ms regions), representing incipient fault conditions. This multi-level severity design with substantial intra-class variance ensures that successful detection requires sophisticated feature extraction beyond magnitude-based approaches.

{While Equation \ref{eq:1} provides a foundational baseline for multi-frequency oscillation and drift, it inherently simplifies the complex, band-specific high-frequency mechanical excitations that occur during real-world arc discharges. Therefore, a gap remains between synthetic and real-world force signals.}

\section{Methodology}

\subsection{Problem Definition}
\noindent\textbf{Multimodal anomaly detection} refers to the task of identifying data instances that exhibit significant deviations from the expected patterns within datasets comprising multiple heterogeneous modalities (e.g., image, force, etc). These modalities may capture complementary aspects of a system or process, and their integration is crucial for robust anomaly identification.
Formally, given a set of aligned multimodal inputs: 
\[
\{(x^{(1)}, x^{(2)}, \dots, x^{(M)})\},
\] 
where each \( x^{(m)} \in \mathcal{X}^{(m)} \) corresponds to the \( m \)-th modality and \( M \) is the total number of modalities, the objective is to learn a model: 
\[
f: \mathcal{X}^{(1)} \times \dots \times \mathcal{X}^{(M)} \rightarrow \{0, 1\},
\] 
that predicts whether a given multimodal instance is {normal} (\( 0 \)) or {anomalous} (\( 1 \)). A subset of the labels for \( x^{(m)}\) can also be provided during training.
The challenge lies in effectively modeling the complex intra- and inter-modal correlations to detect anomalies that may manifest in one or more modalities, while preserving robustness to noise or missing modalities.

\begin{figure}[t!]
 \centering \includegraphics[width=\linewidth]{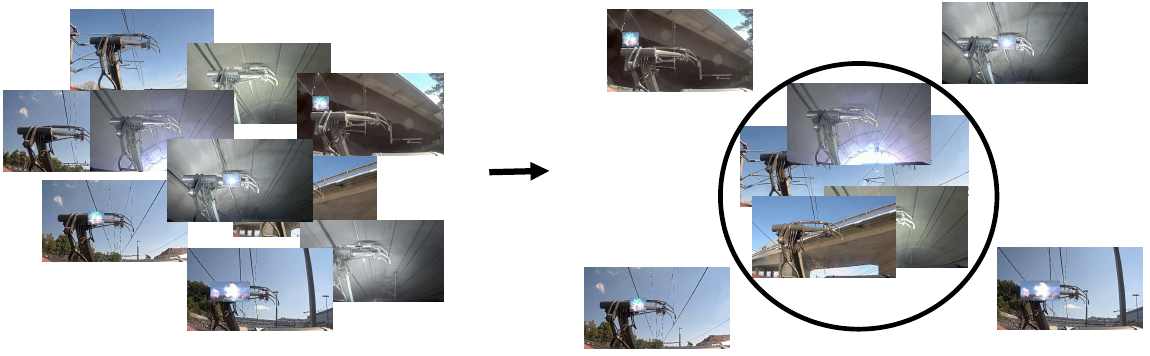}
\vspace{-1.0cm}
\caption{Illustration of the DeepSAD objective: normal samples are embedded close to a hypersphere center, while anomalous samples are pushed outside the hypersphere beyond a margin.}
\label{fig:framework}
\end{figure}

\subsection{Preliminaries}
\noindent\textbf{Deep Semi-Supervised Anomaly Detection (DeepSAD)}~\cite{ruff2019deep} is a deep learning method designed to effectively identify anomalies by leveraging both normal data and a small set of labeled anomalies (\cref{fig:framework}). It operates by training a neural network parameterized by weights $\mathbf{W}$ to learn a mapping function $\phi(\cdot; \mathbf{W})$ from the input space $\mathcal{X} \subseteq \mathbb{R}^d$ to a lower-dimensional feature space $\mathcal{F} \subseteq \mathbb{R}^p$, such that $\mathbf{F} = \phi(x; \mathbf{W})$. The core objective of this transformation is twofold: to map normal data points to be tightly clustered around a pre-defined or learned center $\mathbf{c} \in \mathcal{F}$ within a hypersphere, and simultaneously to ensure that known anomalous data points are mapped outside this hypersphere. This is typically achieved by minimizing an objective function that incorporates these goals. For a set of normal samples $\mathcal{D}_n$ and a set of labeled anomalous samples $\mathcal{D}_a$, the objective function can be generally formulated as:
\begin{equation} \label{eqn:DeepSAD}
\min_{\mathbf{W}} \frac{1}{|\mathcal{D}_n|} \sum_{x_i \in \mathcal{D}_n} ||\phi(x_i; \mathbf{W}) - \mathbf{c}||^2 + \eta \frac{1}{|\mathcal{D}_a|} \sum_{x_j \in \mathcal{D}_a} (||\phi(x_j; \mathbf{W}) - \mathbf{c}||^2)^{-1}, 
\end{equation}
where $|\mathcal{D}_n|$ and $|\mathcal{D}_a|$ denote the number of elements in $\mathcal{D}_n$ and $\mathcal{D}_a$.
Here, the first term {in \cref{eqn:DeepSAD}} minimizes the squared Euclidean distance for normal samples to the center $\mathbf{c}$. The second term, weighted by $\eta > 0$, penalizes labeled anomalies $x_j \in \mathcal{D}_a$ 
and effectively pushes them away from the normal data cluster. 
If no labeled anomalies are available ($|\mathcal{D}_a|=0$), the objective reduces to that of DeepSVDD~\cite{ruff2018deep}, which focuses solely on compacting the normal data. In this case, the objective can be written as:
\begin{equation} \label{eqn:DeepSVDD}
\min_{\mathbf{W}} \frac{1}{|\mathcal{D}_n|} \sum_{x_i \in \mathcal{D}_n} ||\phi(x_i; \mathbf{W}) - \mathbf{c}||^2. 
\end{equation}

During inference, the anomaly score $s(x)$ for a new, unseen sample $x$ is calculated as its squared Euclidean distance to the center $\mathbf{c}$ in the learned feature space: 
\begin{equation} \label{eqn:infer}
s(x) = ||\phi(x; \mathbf{W}^*) - \mathbf{c}||^2,
\end{equation}
where $\mathbf{W}^*$ represents the optimized network weights. A sample $x$ is classified as an anomaly if its score $s(x)$ exceeds a predefined threshold, which is often determined based on the distribution of scores on the training data (e.g., the 95th percentile). A higher score $s(x)$ indicates a greater deviation from the learned normality and thus a higher probability that the sample is an anomaly.

\begin{figure}[t!]
 \centering \includegraphics[width=\linewidth]{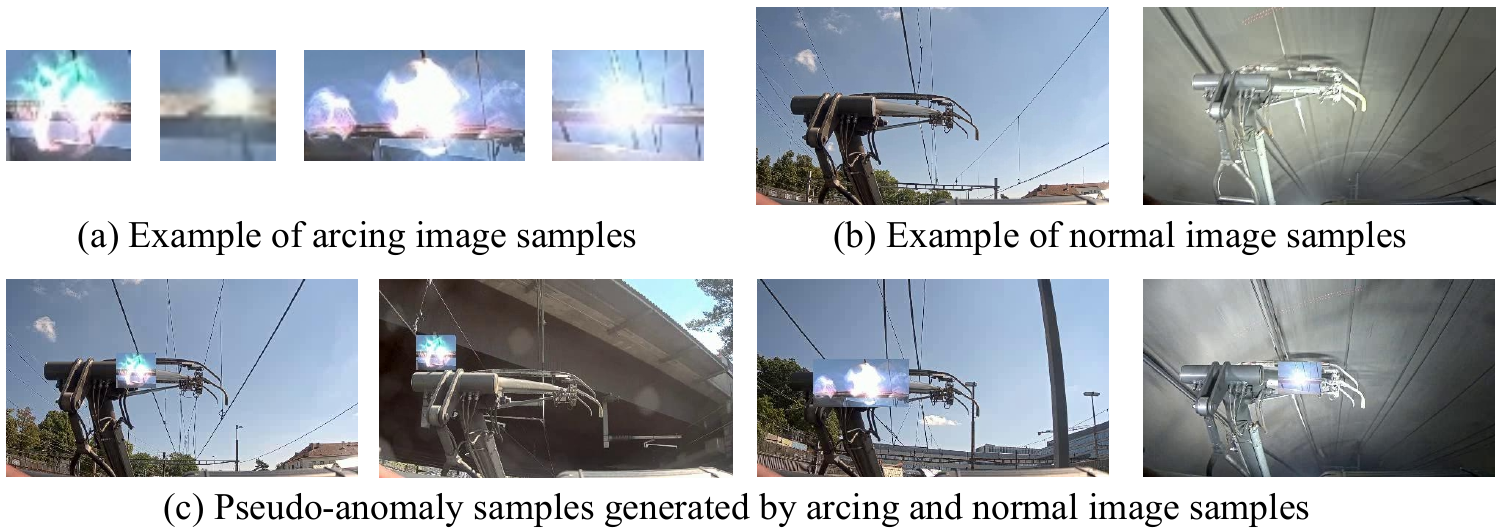}
\vspace{-1.0cm}
\caption{Illustration of pseudo-anomaly image samples generation, where we randomly paste arcing images onto pantographs in normal images. }
\label{fig:ano_img}
\end{figure}

\subsection{Multimodal DeepSAD with Pseudo-anomaly Generation}
In this work, we extend DeepSAD to the multimodal setting and develop effective pseudo-anomaly generation strategies for both image and force modalities. These pseudo-anomalies are then leveraged during training to improve the model’s arcing detection performance.

\begin{figure}[t!]
 \centering \includegraphics[width=\linewidth]{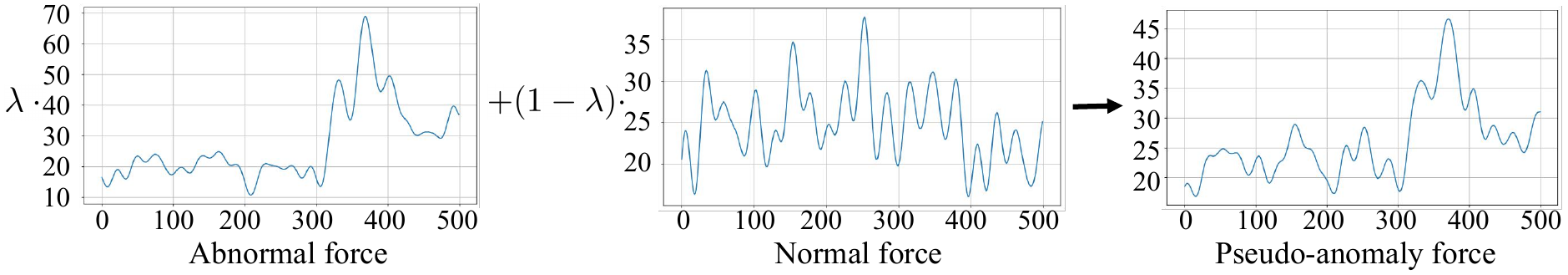}
\vspace{-1.0cm}
\caption{Illustration on pseudo-anomaly force sample generation, where we randomly combine samples from normal and abnormal force patterns. $\lambda \sim \text{Beta}(\alpha, \alpha)$ is a random value between 0 and~1.}
\label{fig:ano_force}
\end{figure}

\noindent\textbf{Pseudo-anomaly Image Sample Generation.} Arcing events are rare and challenging to capture in real-world scenarios. As a result, supervised deep learning models often struggle to learn robust representations only a limited number of labeled anomalies are available. To address this challenge, we propose a pseudo-anomaly generation strategy that leverages a small set of labeled anomalies along together a large set of normal data. We first manually crop the arcing regions in the labeled anomaly samples (\cref{fig:ano_img} (a)). Then, we generate synthetic anomalous image samples by randomly pasting the labeled arcing regions onto pantographs in normal images (\cref{fig:ano_img} (b)), resulting in pseudo-anomalies with simulated arcing (\cref{fig:ano_img} (c)). This approach allows us to generate an unlimited number of anomalous image samples, which are added to the anomaly set $\mathcal{D}_a$ to improve training efficacy.
{This Cut-Paste strategy physically mimics real-world arcing, where intense light emissions are locally superimposed onto the structural background without altering the global geometry of the scene.}

\noindent\textbf{Pseudo-anomaly Force Sample Generation.} Similarly, for the force modality, we assume access to a small set of real anomalous force signals along with a substantial amount of normal samples. Inspired by Mixup~\cite{zhang2018mixup}, we generate synthetic anomalous force signals by taking convex combinations of randomly selected abnormal and normal signals (\cref{fig:ano_force}):
\begin{equation}
 \mathbf{d} = \lambda \mathbf{a_1} + (1 - \lambda) \mathbf{a_2}, 
\label{eq:mixup}
\end{equation}
where $\mathbf{a_1}$ and $\mathbf{a_2}$ represent abnormal and normal force samples, respectively, and $\lambda \sim \text{Beta}(\alpha, \alpha)$, for $\alpha \in (0, \infty)$. This method similarly enables the creation of an arbitrary number of pseudo-anomalous force samples, which are incorporated into $\mathcal{D}_a$.
{Mixup is applied strictly in the time domain before any frequency transformations. This linearly mimics the physical superposition of anomalous transient forces upon standard operational vibrations.}

\noindent\textbf{Multimodal DeepSAD (MultiDeepSAD)}. The original DeepSAD framework is designed for unimodal inputs. We extend it to a multimodal setting by introducing dedicated mapping functions for each modality. Specifically, we define the encoder for each modality as follows: $\phi(\cdot; \mathbf{W}) = [\phi_1(\cdot; \mathbf{W_1}); \phi_2(\cdot; \mathbf{W_2})]$, where $\phi_1(\cdot; \mathbf{W_1})$ and $\phi_2(\cdot; \mathbf{W_2})$ are modality-specific encoders for image and force inputs $x^{(1)}$ and $x^{(2)}$, respectively. The final multimodal representation $\mathbf{F} = \phi(x; \mathbf{W}) = MLP[\mathbf{F_1}; \mathbf{F_2}]$ is the concatenation of features $\mathbf{F_1} = \phi(x_1; \mathbf{W_1})$ and $\mathbf{F_2} = \phi(x_2; \mathbf{W_2})$ from both modalities followed by a multilayer perceptron. The inclusion of modality-specific pseudo-anomalies in $\mathcal{D}_a$ ensures that the model learns more robust and discriminative representations for arcing detection. Finally, we use the same \cref{eqn:infer} as in DeepSAD for inference, but propose \textbf{\textit{a new training objective}} different from \cref{eqn:DeepSAD}:
\begin{equation} \label{eqn:DeepSAD2}
\min_{\mathbf{W}} \frac{1}{|\mathcal{D}_n|} \sum_{x_i \in \mathcal{D}_n} ||\phi(x_i; \mathbf{W}) - \mathbf{c}||^2 + \eta \frac{1}{|\mathcal{D}_a|} \sum_{x_j \in \mathcal{D}_a} \exp(-||\phi(x_j; \mathbf{W}) - \mathbf{c}||^2). 
\end{equation}
Compared with the original DeepSAD anomaly term $(\|\phi(x;W)-\mathbf{c}\|^2)^{-1}$, our exponential formulation $\exp(-\|\phi(x;W)-\mathbf{c}\|^2)$ provides a smoother and numerically more stable optimization objective. Specifically, the original penalty is unbounded when an anomalous sample is mapped close to the center $\mathbf{c}$, which can lead to gradient explosion and high sensitivity to outliers or noisy anomaly labels. In contrast, the exponential term is bounded and yields well-behaved gradients, preventing single hard samples from dominating training and improving robustness. Moreover, the exponential penalty naturally emphasizes ``hard'' anomalies near the decision boundary while rapidly down-weighting anomalies that have already been pushed sufficiently far from $\mathbf{c}$, resulting in a more focused and stable separation between normal and anomalous representations.

{\textit{Mathematical Justification.}
The original DeepSAD anomaly penalty is $(||f(x) - c||^2)^{-1}$. The gradient magnitude is proportional to $||f(x)-c||^{-3}$. As an anomalous sample is pushed close to the center $c$ (e.g., early in training or due to noisy labels), the gradient approaches infinity, leading to numerical instability. In contrast, our proposed penalty is $\exp(-||f(x)-c||^2)$. The gradient is proportional to $-2(f(x)-c)\exp(-||f(x)-c||^2)$. As $f(x) \to c$, the gradient smoothly approaches $0$, preventing gradient explosion and providing stable separation boundaries.}

\section{Experiments}




\renewcommand\arraystretch{1.3}
\begin{table}[t!]
\resizebox{\linewidth}{!}{
\begin{tabular}{l|ccccc}
\hline
    Method & Force& Image&   SBB-AD&   Open-AD  \\

    \hline    
    IForest~\cite{liu2008isolation} 
   &$\checkmark$ & & 60.56& 62.20 \\  
     OCSVM~\cite{scholkopf2001estimating} 
   &$\checkmark$ & & 65.98 & 83.21\\  
    LOF~\cite{breunig2000lof} 
   & $\checkmark$& & 64.63 & 85.25\\  
    KNN~\cite{ramaswamy2000efficient} 
   & $\checkmark$& & 64.98& 84.34 \\  
     AutoEncoder~\cite{hinton2006reducing} 
   & $\checkmark$& & 56.34& 82.13 \\  
     {ArcSE~\cite{quan2024arcse} }
   & &$\checkmark$ &85.24  & 88.78  \\  
     {ArcMSD~\cite{yan2025novel}} 
   & & $\checkmark$&  86.18&   89.66\\  
     DeepSVDD~\cite{ruff2018deep} 
   & $\checkmark$&$\checkmark$ & 68.23&  87.29\\

   DeepSAD~\cite{ruff2019deep} 
   & $\checkmark$&$\checkmark$ & 88.73 &90.04 \\
   
   {Crossmodal Feature Mapping~\cite{costanzino2024multimodal} (adapted) †}
   & $\checkmark$&$\checkmark$ &90.68  &91.75 \\

   MultiDeepSAD* (ours)
   & $\checkmark$&$\checkmark$ & 91.16& 92.41 \\

   MultiDeepSAD (ours)
   & $\checkmark$&$\checkmark$ & 93.49& 95.03 \\

\hline
    \end{tabular}
}
\caption{Results on SBB-AD and Open-AD datasets using different anomaly detection methods. The AUROC is reported. MultiDeepSAD* denotes the variant in which our proposed training objective in \cref{eqn:DeepSAD2} is replaced with the original DeepSAD loss in \cref{eqn:DeepSAD}. {† Crossmodal Feature Mapping~\cite{costanzino2024multimodal} was designed for spatially registered RGB–point-cloud inputs with patch-level feature mapping; we adapt it to the image–force pairing using global-feature mapping. Results should be interpreted as an adapted variant rather than a direct application of the original method.}}
\label{tab:results_DeepSVDD}
\end{table}




\renewcommand\arraystretch{1.3}
\begin{table}[t!]
\resizebox{0.7\linewidth}{!}{
\begin{tabular}{l|ccccc}
\hline
    Method & Force& Image&   SBB-AD&   Open-AD  \\

    \hline    
   MultiDeepSAD
   & $\checkmark$&  &  70.66 & 83.30\\
   MultiDeepSAD
   & &$\checkmark$ & 89.43& 91.64 \\

   MultiDeepSAD
   & $\checkmark$&$\checkmark$ & 93.49& 95.03 \\

\hline
    \end{tabular}
}
\caption{Results on SBB-AD and Open-AD datasets using different modalities.}
\label{tab:results}
\end{table}

\noindent\textbf{Implementation Details.} We use ResNet-18~\cite{he2016deep} as the encoder for images and a two-layer multilayer perceptron (MLP) for force data. The raw force signal is first transformed into the frequency domain using the Fast Fourier Transform (FFT)~\cite{nussbaumer1982fast}, and the first half of the normalized magnitude spectrum is retained as the input feature vector. To initialize the force encoder, we pretrain it using an autoencoder with a reconstruction loss for $100$ epochs. The embeddings from the image and force encoders are then concatenated and passed through an additional two-layer MLP to obtain the final fused representation. We train all networks using the Adam optimizer~\cite{Adam} with a learning rate of $0.0001$. The networks are trained for $30$ epochs with a batch size of $64$, and we set $\eta$ to $1$. Evaluation is performed using the Area Under Receiver Operating Characteristic curve (AUROC) as the standardized metric for all experiments.

\noindent\textbf{Results on SBB-AD and Open-AD Datasets.} \cref{tab:results_DeepSVDD} presents a comparative evaluation of various anomaly detection architectures on the SBB-AD and Open-AD datasets, using AUROC as the performance metric. The results indicate that standard unsupervised methods, including IForest~\cite{liu2008isolation}, OCSVM~\cite{scholkopf2001estimating}, LOF~\cite{breunig2000lof}, KNN~\cite{ramaswamy2000efficient}, AutoEncoder~\cite{hinton2006reducing}, generally achieve lower performance, with AUROC scores often falling below $70$\% on the SBB-AD dataset. DeepSVDD~\cite{ruff2018deep}, which incorporates both force and image data, shows a marked improvement over these classical baselines. Furthermore, DeepSAD~\cite{ruff2019deep} significantly outperforms the unsupervised approaches, demonstrating the substantial benefit of leveraging labeled anomalous data within a semi-supervised setting. Most notably, the proposed MultiDeepSAD method yields the highest detection accuracy on both datasets, achieving AUROC scores of $93.49$\% and $95.03$\% on SBB-AD and Open-AD, respectively, highlighting the clear advantage of our proposed training strategy and pseudo-anomaly generation method for robust arcing anomaly detection. 
{MultiDeepSAD also outperforms recent state-of-the-art multimodal (Crossmodal Feature Mapping~\cite{costanzino2024multimodal}) and vision-based (ArcSE~\cite{quan2024arcse}, ArcMSD~\cite{yan2025novel}) baselines.}
{For all baselines, the image and force encoders share the same architecture as those used in our framework. For Crossmodal Feature Mapping, we follow the paper and employ two lightweight mapping networks trained exclusively on normal data: $f_{img \rightarrow force}$ and $f_{force \rightarrow img}$. Both networks are implemented as three-layer MLPs, each but the last one followed by GeLU activations~\cite{costanzino2024multimodal}. Because patch correspondences do not exist in our setting, the mappings operate on the global feature vectors, and the anomaly score is the sum of the two L2 discrepancies between mapped and observed features, replacing the original patch-level anomaly map and its spatial aggregation.}

\cref{tab:results} analyzes the impact of input modalities on MultiDeepSAD performance. Using force data alone yields relatively limited accuracy ($70.66$\% AUROC on SBB-AD and $83.30$\% on Open-AD), whereas using image data alone leads to a substantial improvement ($89.43$\% and $91.64$\% AUROC, respectively), indicating that visual cues provide strong discriminative signals for arcing detection. Notably, combining force and image modalities achieves the best results on both datasets ($93.49$\% on SBB-AD and $95.03$\% on Open-AD), demonstrating that the two modalities offer complementary information and that multimodal fusion consistently enhances robustness and overall detection performance.

\begin{figure}[t!]
 \centering \includegraphics[width=\linewidth]{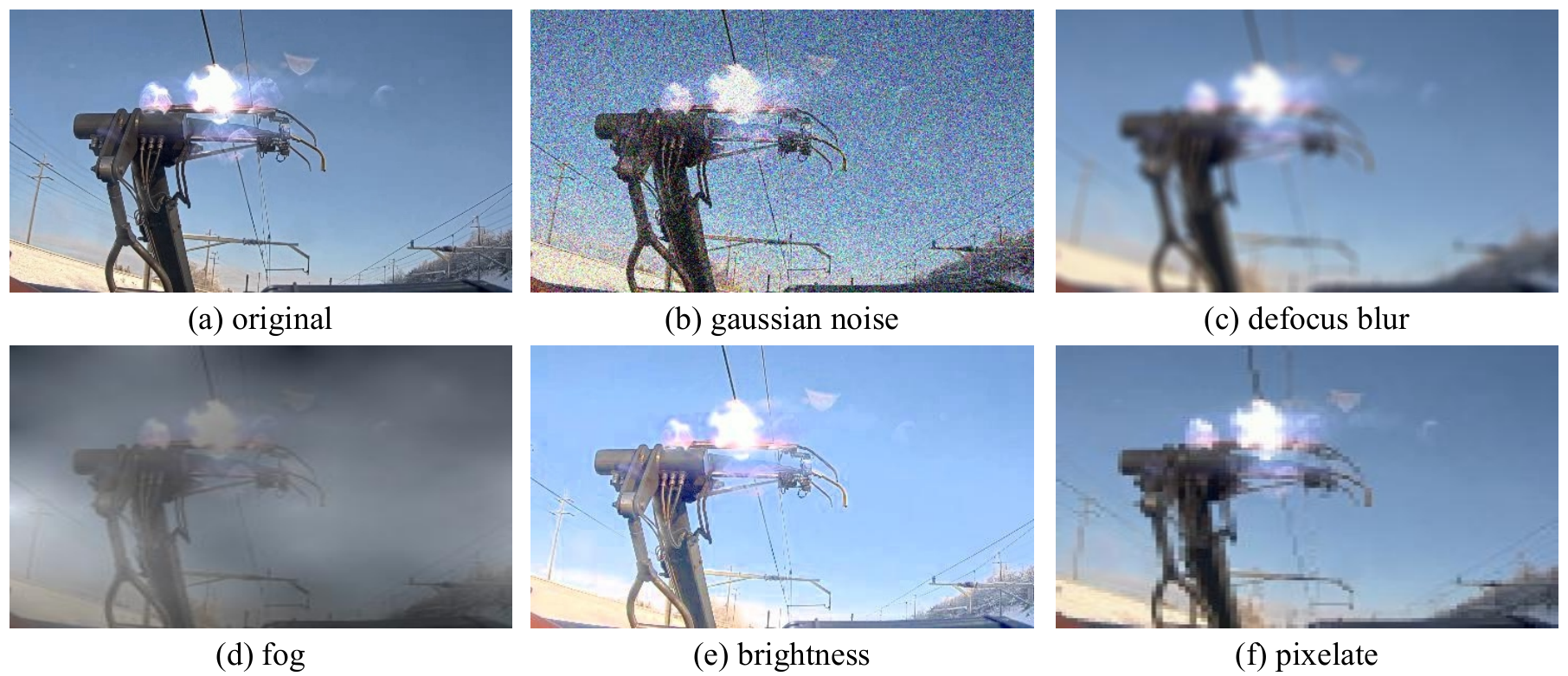}
\vspace{-1.0cm}
\caption{Illustration of different corruptions on the original image (level 3). }
\label{fig:corrupt}
\end{figure}

\noindent\textbf{Robustness to Corruptions on Images.} In real-world settings, environmental conditions may vary significantly (e.g., changing from sunny to cloudy or rainy)~\cite{dong2025advances}, posing challenges for consistent arcing detection. Such variations often manifest as visual distortions due to lighting changes, lens contamination, or motion-induced blur in on-board monitoring systems. To simulate such distribution shifts, we evaluate our model under several common corruptions, including Gaussian noise, defocus blur, fog, brightness change, and pixelation, applied only to the image modality during testing (\cref{fig:corrupt}). These perturbations are selected to reflect plausible degradations in pantograph-catenary monitoring, such as reduced visibility due to weather (fog), camera defocus from vibrations, sensor noise in low-light conditions, or compression artifacts in video transmission. The model is trained on clean images using both modalities. As shown in \cref{tab:corrupt_l1}, our framework maintains strong performance across most corruptions on SBB-AD. Among them, ‘fog’ has the most significant adverse impact, while other corruptions cause only minor degradation, demonstrating the model’s robustness to a wide range of visual distortions.

\renewcommand\arraystretch{1.3}
\begin{table}[t!]
\resizebox{0.7\linewidth}{!}{
\begin{tabular}{l|ccc}
\hline
    Corruption & AUROC (level 1)& AUROC (level 3)\\

    \hline    
    Gaussian noise
   &  91.07& 89.25  \\
    Defocus blur
   &  87.95&  88.05 \\
    Fog
   &  77.83&  67.08 \\
    Brightness
   &   92.37& 87.88 \\
    Pixelate
   & 88.89&  88.28 \\

\hline
    \end{tabular}
}
\caption{Results on different corruptions at different levels.}
\label{tab:corrupt_l1}
\end{table}

{\noindent\textbf{Cross-Dataset Generalization.} We conduct a cross-dataset evaluation where the model is trained on Open-AD and tested on SBB-AD. While performance naturally drops due to the domain shift (achieving an AUROC of 81.4\% compared to the in-domain 93.49\%), it remains highly competitive. The primary domain gap arises because our synthetic force signals simplify the high-frequency structural resonances and non-linear aerodynamic damping effects present in the physical pantograph system of the SBB dataset. Importantly, this evaluation demonstrates the practical value of the proposed approach, as it assesses whether a model trained on an openly available dataset can be transferred to previously unseen proprietary datasets without requiring access to labeled data from the target domain. This setting closely reflects real-world deployment scenarios, where organizations often seek to leverage open datasets to develop models that can subsequently be applied to their own operational data. In future work, a small number of real samples could be leveraged for fine-tuning to further bridge the domain gap.}

\noindent\textbf{Robustness to the Number of Real Anomaly Samples.} \cref{tab:num} evaluates the sensitivity of our method to the number of real anomaly samples used for pseudo-anomaly generation on the SBB-AD dataset. With only one real anomaly, the model already achieves a strong AUROC of $84.34$\%, demonstrating effectiveness in a highly low-shot setting. Increasing the number of real anomalies to two yields a substantial performance jump to $93.34$\% AUROC, after which the results remain consistently high and stable, ranging from $92.16$\% to $93.49$\% for three to five anomalies. Overall, the proposed pseudo-anomaly generation is robust to the choice of anomaly count and can reach near-saturated performance with only a few labeled anomalies. These results highlight the practicality of our method in scenarios where faults are extremely rare and difficult to obtain.





\renewcommand\arraystretch{1.3}
\begin{table}[t!]
\resizebox{0.8\linewidth}{!}{
\begin{tabular}{l|ccccc}
\hline
    Number of real anomalies & 1 & 2& 3& 4& 5\\

    \hline    
    AUROC
   & 84.34  &  93.34  &  92.16 & 92.95  & 93.49  \\

\hline
    \end{tabular}
}
\caption{Robustness to different numbers of real anomaly samples for pseudo-anomaly generation.}
\label{tab:num}
\end{table}

\noindent\textbf{Ablation on Force Data Representations.} The raw force signal is a one-dimensional time-series sequence that can be transformed into different representations, such as the Wavelet transform~\cite{zhang2019wavelet}, short-time Fourier transform (STFT)~\cite{durak2003short}, Gramian Angular Fields (GAF), Markov Transition Fields (MTF), and Recurrence Plots (RP)~\cite{wang2015encoding}, and FFT. Overall, performance varies noticeably across feature transformations on SBB-AD: time-frequency encoding tends to be more effective than raw signals or some image-like mappings. As shown in \cref{tab:trans}, FFT achieves the best AUROC ($70.66$\%), followed by RP ($66.24$\%), while Raw and STFT provide comparable mid-range results ($65.66$\% and $65.27$\%). Wavelet performs slightly worse ($64.22$\%), and GAF/MTF yield the lowest AUROC ($61.00$\% and $61.86$\%). These results suggest that, for force data, frequency-domain features are most informative for distinguishing arcing anomalies, whereas certain image-based time-series transforms may discard discriminative structure in this setting. Based on these findings, we use FFT as the default representation for force data in all experiments.




\renewcommand\arraystretch{1.3}
\begin{table}[t!]
\resizebox{0.8\linewidth}{!}{
\begin{tabular}{l|ccccccc}
\hline
    Method & Raw& Wavelet&STFT &GAF &MTF &RP &  FFT\\

    \hline    
    AUROC
   & 65.66 & 64.22 & 65.27& 61.00& 61.86& 66.24  & 70.66 \\

\hline
    \end{tabular}
}
\caption{Ablation on different representations for force data.}
\label{tab:trans}
\end{table}

\noindent\textbf{Comparison to Alternative Pseudo-anomaly Generation Approaches.}
We compare our pseudo-anomaly generation strategy with alternative approaches on SBB-AD, including Mixup~\cite{zhang2018mixup} and NNG-Mix~\cite{dong2023nngmix}. Mixup generates pseudo-anomalies by interpolating between anomaly and normal samples, whereas NNG-Mix uses only the nearest normal neighbors of anomaly samples for mixing. These comparison approaches were selected as representative strategies for data augmentation and semi-supervised anomaly detection, with Mixup being a widely adopted method in classification, and NNG-Mix offering a more structure-aware approach tailored to anomaly contexts. These methods were originally designed for unimodal scenarios. We extend them to multimodal settings by applying the same operations independently to each modality. As shown in \cref{tab:gen}, our method consistently outperforms both alternatives, highlighting the effectiveness of our modality-aware and targeted pseudo-anomaly generation approach when used with Multimodal DeepSAD. Interestingly, training with Mixup samples actually decreases performance. This may be due to the fact that Mixup’s image mixing strategy also blends the backgrounds, potentially diminishing the relevance of the generated anomalies for arcing detection.




\renewcommand\arraystretch{1.3}
\begin{table}[t!]
\resizebox{0.8\linewidth}{!}{
\begin{tabular}{l|cccc}
\hline
    Method & w/o anomaly generation & Mixup&NNG-Mix &  Ours\\

    \hline   
   AUROC
    & 90.31&  90.23 &  92.08&  93.49\\

\hline
    \end{tabular}
}
\caption{Ablation on different anomaly generation methods.}
\label{tab:gen}
\end{table}







\noindent\textbf{Ablation on Multimodal Fusion Methods.} \cref{tab:fuse} compares different multimodal fusion strategies on SBB-AD.
Late fusion concatenates image and force embeddings and feeds them to an MLP to learn cross-modal interactions. Weighted fusion forms a convex combination of modality embeddings using learnable scalar weights. Gated fusion~\cite{arevalo2017gated} applies learnable, per-dimension gates to modulate each modality’s contribution, optionally with modality dropout for robustness. Attention fusion~\cite{vaswani2017attention} computes sample-wise modality weights via a softmax-based attention mechanism.
Late fusion achieves the best performance with $93.49$\% AUROC, outperforming more complex alternatives including weighted fusion ($91.56$\%), gated fusion ($91.61$\%), and attention-based fusion ($91.84$\%). This suggests that, in our setting, a straightforward late fusion of modality-specific features is sufficient to effectively exploit cross-modal complementarity, while additional learnable fusion mechanisms do not provide further gains and may introduce unnecessary optimization complexity or overfitting.
{The superior performance of Late Fusion is likely due to the highly distinct and already-discriminative nature of the representations extracted from the two modalities. In pantograph arcing, the visual signature (bright flash) and mechanical signature (sudden force transient) are somewhat independent physical phenomena occurring simultaneously. Complex attention or cross-modal gating mechanisms tend to overcomplicate the learning objective and are prone to overfitting on datasets of limited scale, whereas Late Fusion robustly aggregates the independent discriminative power of both branches.}

\renewcommand\arraystretch{1.3}
\begin{table}[t!]
\resizebox{0.6\linewidth}{!}{
\begin{tabular}{l|cccc}
\hline
    Method & Late& Weighted& Gated &  Attention\\

    \hline   
   AUROC
    & 93.49&  91.56 &  91.61&  91.84\\

\hline
    \end{tabular}
}
\caption{Ablation on multimodal fusion methods.}
\label{tab:fuse}
\end{table}

\noindent\textbf{Ablation on Different Image Backbones.} \cref{tab:back} studies the effect of different image backbones on SBB-AD. ResNet-18 achieves the best performance with $89.43$\% AUROC, slightly outperforming ResNet-50 ($88.89$\%) and ViT-B/16 ($88.08$\%)~\cite{dosovitskiy2020image}. Overall, the results indicate that detection accuracy is relatively insensitive to the choice of backbone within this range, and that a lightweight CNN backbone (ResNet-18) is sufficient to provide strong visual representations for our task.

\renewcommand\arraystretch{1.3}
\begin{table}[t!]
\resizebox{0.6\linewidth}{!}{
\begin{tabular}{l|ccc}
\hline
    Model & ResNet-18& ResNet-50& ViT-B/16\\

    \hline   
   AUROC
    & 89.43 &  88.89  & 88.08\\

\hline
    \end{tabular}
}
\caption{Ablation on different image backbones.}
\label{tab:back}
\end{table}

\renewcommand\arraystretch{1.3}
\begin{table}[t!]
\resizebox{0.65\linewidth}{!}{
\begin{tabular}{lcccc}
\toprule
{Modality} & {0-shot} & {2-shot} & {4-shot} & {8-shot} \\
\midrule
\multicolumn{5}{l}{\textit{Image}} \\
RGB Images & 77.37 & {86.89} & 83.07 & 81.28 \\
\midrule
\multicolumn{5}{l}{\textit{Force}} \\
Raw & 46.09 &  {49.48} & 48.97 & 48.57 \\
RP            & 42.32 & 45.21 & 43.91 &  {45.79} \\
GAF           &  {55.27} & 53.84 & 51.99 & 51.86 \\
MTF           & {46.97} & 37.89 & 38.48 & 37.26 \\
\midrule
\multicolumn{5}{l}{\textit{Multimodal Fusion}} \\
RGB + GAF& 74.05 & 73.91 & {83.80} & 76.62 \\
\bottomrule
\end{tabular}
}
\caption{Comparison to multimodal large language models.}
\label{tab:mllm}
\end{table}

\noindent\textbf{Comparison to Multimodal Large Language Models (MLLMs).} MLLMs have demonstrated strong performance across a wide range of multimodal tasks involving text, images, videos, and audio. However, their effectiveness in industrial safety monitoring, such as railway arcing detection, remains underexplored. In this study, we investigate whether a general-purpose MLLM can serve as a viable alternative to traditional arcing detection methods. Specifically, we evaluate Qwen3-VL-2B-Instruct~\cite{Qwen3-VL} and formulate arcing detection as a prompted binary classification task. The system prompt describes the visual characteristics of electrical arcing (e.g., sparks or flashes between the pantograph and the overhead contact wire) and constrains the model to output a binary decision. In the few-shot setting, we provide $k$ annotated example images to calibrate the model to dataset-specific visual patterns. As shown in \cref{tab:mllm}, the model achieves strong performance on RGB images on SBB-AD, indicating that Qwen3-VL has robust visual understanding for identifying electrical arcing. In contrast, it fails to meaningfully interpret force-based signal visualizations (Raw plots, RP, GAF, MTF), yielding near-random AUROC values, which suggests limited prior knowledge for these representations. Overall, Qwen3-VL performs substantially worse than our proposed MultiDeepSAD.

\begin{figure}[t!]
 \centering \includegraphics[width=\linewidth]{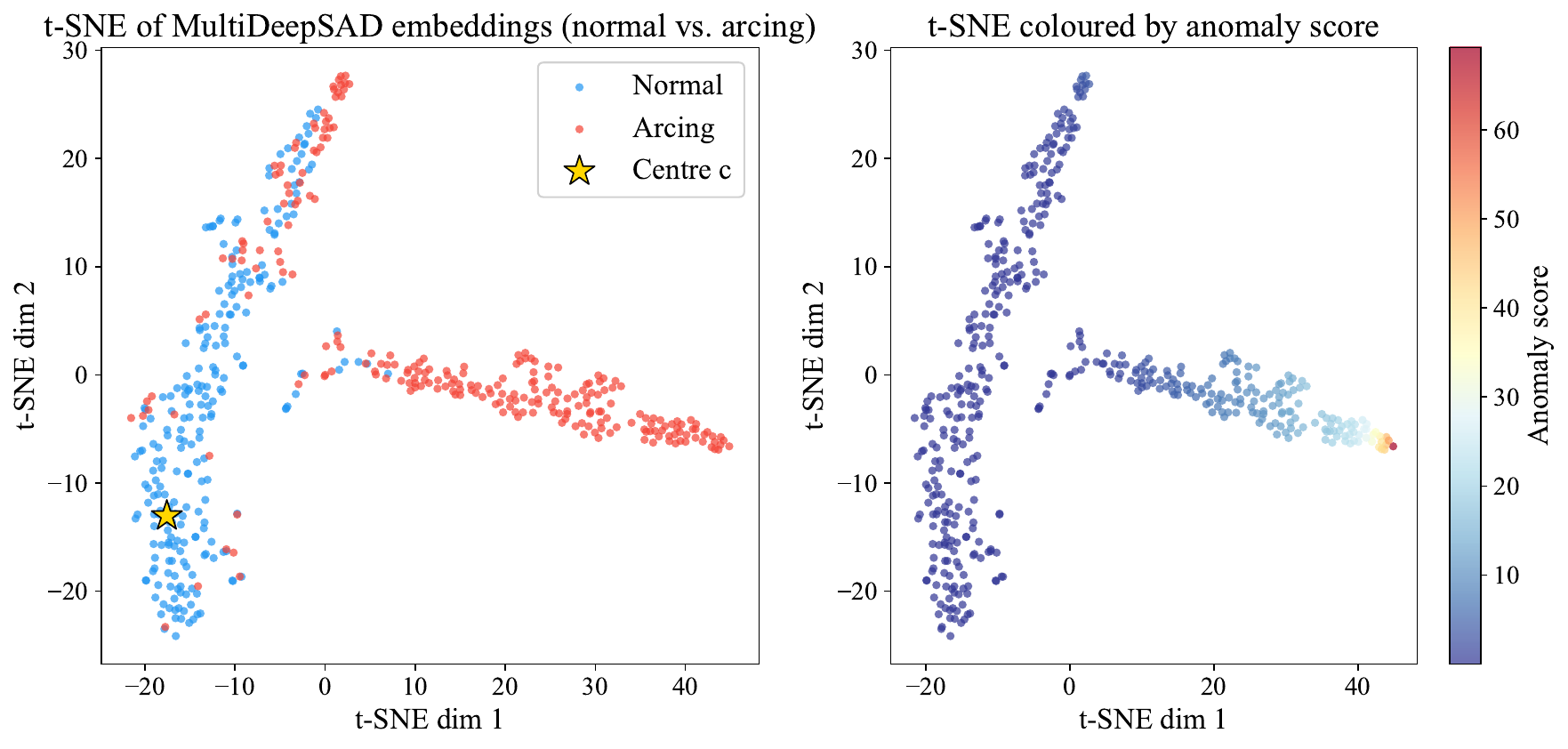}
\vspace{-1.2cm}
\caption{{t-SNE visualizations of the latent feature space for normal and arcing samples. }}
\label{fig:tsne}
\end{figure}

{\noindent\textbf{Visualizations of the Feature Space.} We generate t-SNE visualizations of the latent feature space using our proposed MultiDeepSAD. The visualization in \cref{fig:tsne} clearly shows that MultiDeepSAD achieves a tight clustering of normal samples and a clear and distinct margin of separation for the anomalous samples.
}

\begin{figure}[t!]
 \centering \includegraphics[width=\linewidth]{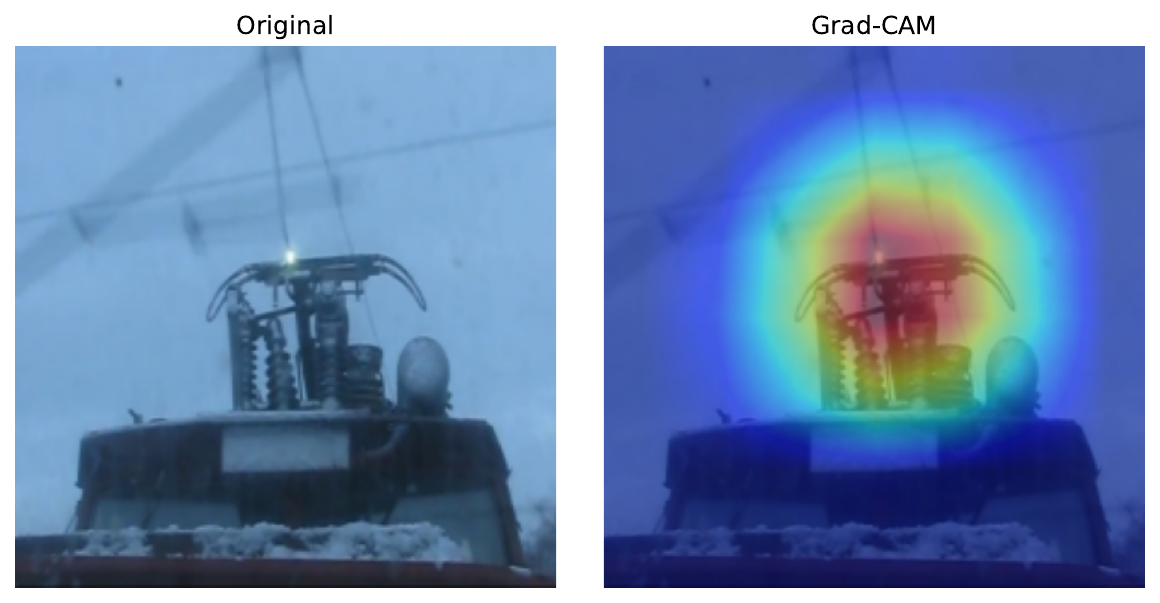}
\vspace{-1.2cm}
\caption{{Grad-CAM visualizations. }}
\label{fig:grad}
\end{figure}

{\noindent\textbf{Visualizations of Grad-CAM.} We add Grad-CAM~\cite{selvaraju2017grad} visualizations (\cref{fig:grad}), with the gradient signal computed with respect to the anomaly score (squared distance to the MultiDeepSAD hypersphere centre) rather than a class logit, so the heatmaps directly reveal which image regions drive the model's anomaly decision. The activation maps consistently concentrate on the bright luminous arc region along the contact wire, rather than on the peripheral edges introduced by the image compositing procedure. This provides direct visual evidence that the model has learned to respond to the physical characteristics of the arc discharge (its luminance, spatial extent, and position relative to the catenary) rather than to synthetic boundary artefacts.
}

{\noindent\textbf{Hyperparameter Sensitivity.} The hyperparameter $\eta$ determines the balance between clustering normal data and repelling anomalies. We have conducted a sensitivity analysis testing $\eta \in \{0.1, 0.5, 1.0, 2.0\}$. The results in \cref{tab:hyper} show that performance is highly stable across this range (AUROC varies between 93.60\% and 95.03\%), with $\eta=1.0$ yielding the optimal balance.}

\renewcommand\arraystretch{1.3}
\begin{table}[t!]
\resizebox{0.6\linewidth}{!}{
\begin{tabular}{l|cccc}
\hline
    $\eta$ & 0.1& 0.5& 1.0& 2.0\\

    \hline   
   AUROC
    & 93.60 &  94.79  & 95.03& 93.67\\

\hline
    \end{tabular}
}
\caption{{Ablation on hyperparameter sensitivity.}}
\label{tab:hyper}
\end{table}

{\noindent\textbf{Results under Different Evaluation Protocols.} We define a causal protocol that treats every one-second window identically, regardless of its label. Each one-second window is decoded at a fixed temporal sampling rate of [K] frames per window (we report [K = 60], i.e., all frames at the native 60 Hz camera rate, and a subsampled variant [K = 20]), applied identically to normal and anomalous windows. The one-second force segment is preprocessed once per window (FFT of the 500-sample signal; first half of the normalized magnitude spectrum), exactly as in training. Each sampled frame is paired with the window's force spectrum and passed through the trained MultiDeepSAD model to obtain a per-frame anomaly score. No component of this pipeline requires arc labels or manual intervention. The window-level anomaly score is obtained by score-level aggregation across the [K] per-frame scores. We report three standard aggregation rules: maximum, mean, and top-[k] mean ([k = 3]). The maximum rule matches the transient physics of arcing — an arc need only be visible in a small number of frames within the window — while the top-[k] mean additionally damps single-frame outliers caused by, e.g., glare, tunnel transitions, or motion blur. The new results on SBB-AD are summarized in Table \ref{tab:causal_protocol}. The single fixed-frame rule degrades performance considerably (72.51 AUROC), because arcing events are visible in only a small fraction of the 60 frames within a window; this confirms that a single-frame protocol is not an adequate basis for deployment claims, and it motivates multi-frame aggregation as the appropriate causal protocol. Second, under multi-frame aggregation, MultiDeepSAD attains 92.12 AUROC (max, K = 60), i.e., 1.37 points below the oracle-frame upper bound, quantifying the optimism introduced by the previous protocol.}

\begin{table}[t]
\centering
\caption{{AUROC (\%) on SBB-AD under different evaluation protocols with MultiDeepSAD.}}
\label{tab:causal_protocol}
\resizebox{1.0\linewidth}{!}{
\begin{tabular}{llc}
\toprule
\textbf{Protocol} & \textbf{Aggregation} & \textbf{AUROC} \\
\midrule
Oracle frame (previous, upper bound) & --- & 93.49 \\
Causal, fixed central frame          & --- & 72.51 \\
Causal, $K=20$ & max / mean / top-3 & 90.08 / 82.16 / 88.98 \\
Causal, $K=60$ & max / mean / top-3 & 92.12 / 85.31 / 91.88 \\
\bottomrule
\end{tabular}
}
\end{table}

{\noindent\textbf{Computational Complexity and Real-Time Applicability.} The framework is highly efficient, largely due to the use of a lightweight ResNet-18 for image data and an MLP for force data. The total parameter count is 11.5M, requiring only 1.82 GFLOPs. On NVIDIA RTX 5060 and AMD Ryzen 9 CPU, the model achieves an inference latency of approximately 1.361 ms per sample (735 FPS), which is more than 24× faster than real-time video (30 FPS), confirming its suitability for online monitoring scenarios. In practice, maintenance decisions are typically not taken on a frame-by-frame basis and instead rely on aggregating evidence over longer temporal windows. Consequently, the proposed architecture substantially exceeds the latency requirements of operational workflows, and real-time deployment would be readily achievable if desired. These results demonstrate that the proposed multimodal anomaly detection framework satisfies the low-latency requirements of real-time railway contact wire monitoring without any model compression or hardware-specific optimization, leaving room for further acceleration (e.g., quantization or pruning) if deployment on edge hardware is required.}

\section{Conclusion}
In this work, we propose a novel framework for detecting electrical arcing in pantograph-catenary systems by integrating image and force data. 
Our proposed MultiDeepSAD architecture extends semi-supervised anomaly detection into the multimodal domain, using pseudo-anomaly generation to make effective use of limited real anomaly data during training. Experimental results demonstrate that our multimodal approach achieves superior detection accuracy and robustness compared to baseline methods, even in the presence of domain shifts or when only a few real arcing samples are available. Ablation studies confirm the complementary value of both visual and force features for arcing detection. 
We also release a new multimodal dataset that includes synchronized visual and force measurements, which can serve as a challenging benchmark for future research on arcing detection in pantograph-catenary systems.
From a practical standpoint, this research demonstrates that combining non-contact visual monitoring with force-based measurements enables a more reliable system for detecting arcing events. Adopting such a multimodal detection strategy has the potential to lower maintenance costs, prevent infrastructure damage, and enhance the operational reliability of rail networks. However, arcing detection alone is not sufficient to inform maintenance actions, as infrastructure risk is driven by the cumulative effect of repeated events. A practical maintenance workflow therefore requires an additional layer that consistently localizes and tracks arcing occurrences over time (i.e., linking detections to specific catenary locations and to individual pantographs) so that event frequency, duration, and severity can be aggregated into a damage or health index. Future work will explore real-time deployment on operational trains, integrate robust spatiotemporal tracking and alignment across runs, and develop lightweight cumulative damage modeling for use on resource-constrained embedded platforms. {Besides, exploring Poisson blending to further improve image synthesis quality is also an interesting direction for future work.}

\section*{Acknowledgment of AI Assistance in Manuscript Preparation}
During the preparation of this work, the authors used ChatGPT to assist with refining and correcting the text. After using this tool, the authors carefully reviewed and edited the content as needed and take full responsibility for the content of this publication.

\section*{Acknowledgment of project funding}
The authors acknowledge the support of "In-service diagnostics of the catenary/pantograph and wheelset axle systems through intelligent algorithms" (SENTINEL) project, supported by the ETH Mobility Initiative.



\color{black}
\bibliographystyle{unsrt}  
\bibliography{PHME_Latex_Template}  

\end{document}